\documentclass[runningheads]{llncs}
\usepackage[T1]{fontenc}
\usepackage{graphicx}
\usepackage{booktabs}
\usepackage[misc]{ifsym}

\usepackage{amsmath}
\usepackage{amssymb}
\usepackage{bm}
\usepackage{bbm}
\usepackage{algorithm}
\usepackage{algpseudocode}
\usepackage{booktabs}
\usepackage{subcaption}
\DeclareMathOperator*{\argmax}{arg\,max}

\usepackage{xcolor,colortbl}
\usepackage{hyperref}
\usepackage[misc]{ifsym}

\begin{document}

\title{Novel Node Category Detection Under Subpopulation Shift}


\author{Hsing-Huan Chung \inst{1}\textsuperscript{(\Letter)} \and 
Shravan Chaudhari\inst{2} \and
Yoav Wald\inst{3} \and
Xing Han\inst{2} \and
Joydeep Ghosh\inst{1}
}

\tocauthor{Hsing-Huan Chung, Shravan Chaudhari, Yoav Wald, Xing Han, Joydeep Ghosh}
\toctitle{Novel Node Category Detection Under Subpopulation Shift}

\authorrunning{H.H. Chung et al.}


\institute{Department of Electrical and Computer Engineering, The University of Texas at Austin, Austin TX, USA
\and
Department of Computer Science, Johns Hopkins University, Baltimore MD, USA
\and
Center for Data Science, New York University, New York NY, USA \email{hhchung@utexas.edu}}

\maketitle              

\begin{abstract}
In real-world graph data, distribution shifts can manifest in various ways, such as the emergence of new categories and changes in the relative proportions of existing categories.
It is often important to detect nodes of novel categories under such distribution shifts for safety or insight discovery purposes.
We introduce a new approach, Recall-Constrained Optimization with Selective Link Prediction (RECO-SLIP), to detect nodes belonging to novel categories in attributed graphs under subpopulation shifts.
By integrating a recall-constrained learning framework with a sample-efficient link prediction mechanism, RECO-SLIP addresses the dual challenges of resilience against subpopulation shifts and the effective exploitation of graph structure.
Our extensive empirical evaluation across multiple graph datasets demonstrates the superior performance of RECO-SLIP over existing methods.
The experimental code is available at: \href{https://github.com/hsinghuan/novel-node-category-detection}{https://github.com/hsinghuan/novel-node-category-detection}.

\end{abstract}

\section{Introduction}

\begin{figure}[t]
\centering
\includegraphics[width=0.9\textwidth]{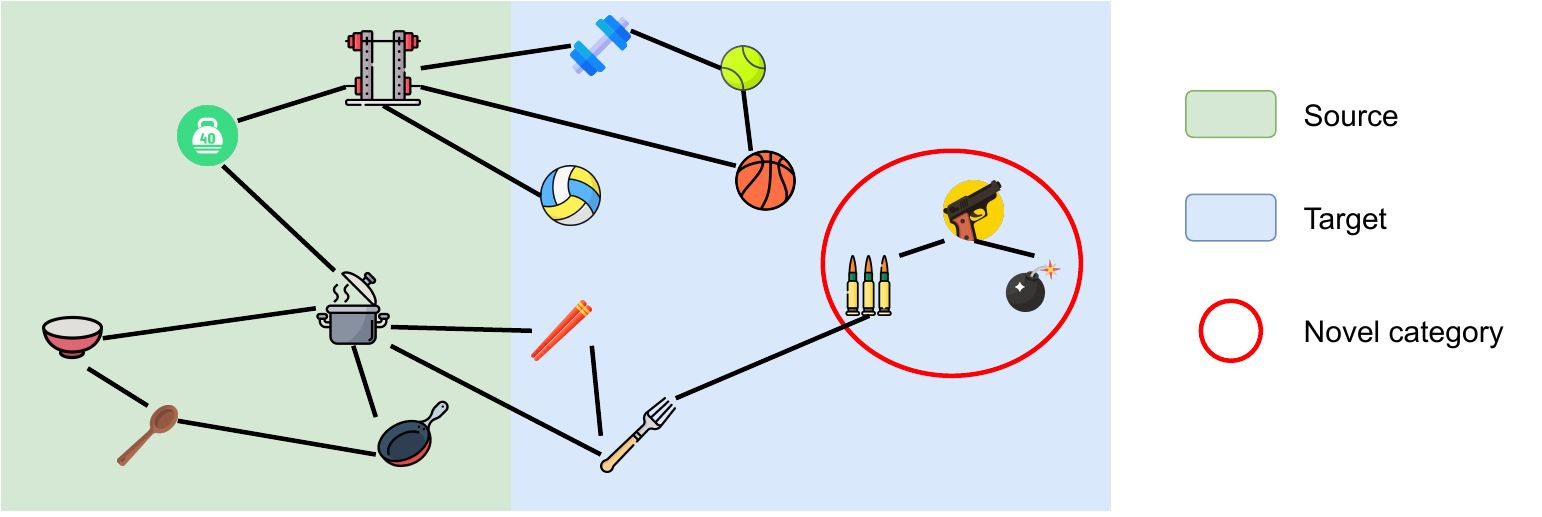}
\caption{An illustration of novel node category detection under subpopulation shift using a product co-purchasing network.
The target domain consists of products from the two categories that exist in the source domain (sports and kitchen) and a novel category (weapons).
Meanwhile, the relative proportion of the two original categories changes from source to target.
The goal is to detect the products belonging to the novel category in the target domain.}
\label{fig:problem}
\end{figure}

Distribution shifts may occur in real-world graphs either through natural temporal evolution \cite{pmlr-v232-chung23a} or the manual integration of new data sources \cite{wu2020unsupervised}.
Such shifts manifest in various ways, including the emergence of new categories or alterations in the relative proportions of existing ones.
The task of novel category detection \cite{blanchard2010semi,wald2023birds} involves identifying data samples that do not fit into pre-existing categories even in the face of these shifts.
As a motivation, consider the product co-purchasing network of a consumer-to-consumer (c2c) e-commerce platform in Fig. \ref{fig:problem}.
The co-purchasing network may evolve due to the e-commerce platform being available in new regions.
The evolution can lead to the introduction of new product categories and changes in the relative popularity of existing ones. 
It is crucial to identify products in potential novel categories due to either safety reasons or the need to develop new insights for platform improvement.
Additionally, the detection has to be able to be done under the subpopulation shift among existing (non-novel) categories, ideally without knowing the category labels of individual products due to the chaotic nature of a c2c platform.
Similar phenomena and demands also exist in academic citation graphs where papers demonstrating new research topics are to be detected.
These applications underscore the importance of studying novel node category detection amidst subpopulation shifts.

In this study, we consider an attributed graph where nodes belong to either the source or the target domain.
The source nodes all fall into the non-novel categories whereas the target nodes may belong to the non-novel categories or a novel category.\footnote{This formulation can be readily extended to multiple novel categories by viewing them as a single category.
The detected novel category can be subsequently partitioned into multiple sub-categories by a suitable graph partitioning algorithm.}
Using the co-purchasing network example, one can think of the source nodes as the products that have already existed before a certain time point and the target nodes as the products introduced afterward.
Our primary goal is to identify nodes within the target domain that fall into the novel category while the relative proportions of non-novel categories between the source and target domains may vary.
As shown by previous work \cite{blanchard2010semi,garg2022domain,wald2023birds}, novel category detection can be naturally reframed as a positive-unlabeled (PU) learning problem \cite{bekker2020learning}.
In this context, non-novel categories are collectively considered as the positive class, and the novel category as the negative class.
Since all source nodes belong to the non-novel categories and the novel/non-novel labels of individual target nodes are not known, the source and target nodes can be viewed as positively labeled and unlabeled data, respectively.
Consequently, the problem of novel node category detection is equivalent to learning from positively labeled and unlabeled nodes.


Despite the connection to PU learning, existing standard and graph PU learning methods encounter significant limitations. 
These methods often rely on the Selected Completely At Random (SCAR) assumption \cite{elkan2008learning}, which posits that each positive instance is equally likely to be labeled.
However, this assumption breaks in the face of subpopulation shifts, leading to compromised PU learning performance.
Several approaches have sought to relax the SCAR assumption.
For instance, propensity weighting-based methods \cite{bekker2019beyond,gerych2022recovering,wang2024pue} estimate the probability of each positive sample being labeled and integrate the probabilities into the loss function.
One other recent method that loosens the SCAR assumption, CoNoC \cite{wald2023birds}, solves a constrained learning problem and has finite sample guarantees under distribution shift. 
Although these methods do not require SCAR to hold, they essentially treat all unlabeled samples as equally negative, or in our language, all target samples as equally novel.
They do not explicitly utilize the subgroup structure provided by edges in a graph to further distinguish nodes of the novel category from the rest.

To address the challenges of vulnerability to subpopulation shifts and the ineffective use of graph structure, we introduce \textbf{RE}call-\textbf{C}onstrained \textbf{O}ptimization with \textbf{S}elective \textbf{LI}nk \textbf{P}rediction (RECO-SLIP).
RECO-SLIP builds upon the constrained learning framework of CoNoC, which has only shown empirical success on tabular datasets.
RECO-SLIP enhances this framework by integrating a link prediction loss induced by a sample-efficient edge sampling strategy to preserve the novel subgroup structure in the node representation space.
Our comprehensive experiments showcase the effectiveness of RECO-SLIP.
In summary, our key contributions are threefold:
\begin{itemize}
    \item We formally define the problem of detecting nodes from novel categories in attributed graphs, particularly under conditions of subpopulation shift.
    \item We introduce RECO-SLIP, which synergizes a recall-constrained learning framework with a sample-efficient link prediction mechanism. This approach addresses the limitations of existing methods under subpopulation shifts and the underutilization of graph structures.
    \item We conduct a comprehensive empirical evaluation of our approach, comparing its performance against standard PU learning, propensity-weighting, and graph PU learning methods on five graph datasets. Our findings affirm the effectiveness and robustness of the proposed solution.
\end{itemize}

\section{Related Work}
\subsection{PU Learning and Novel Category Detection}
PU learning \cite{bekker2020learning} is the problem of learning from a dataset with only positively labeled and unlabeled data.
Novel category detection \cite{blanchard2010semi,wald2023birds} can be naturally reframed as a PU learning problem. 
The concept of novel and non-novel in our problem can be mapped to the concept of negative and positive in PU learning, respectively.
Then, the role of the source nodes would correspond to the positively labeled samples in PU learning since all source nodes are known to be from the non-novel (positive) categories.
The target nodes would correspond to the unlabeled samples because they could be from a novel (negative) or non-novel (positive) category and the learner does not know which ones are novel and which ones are not.
This connection has been drawn by several prior work \cite{blanchard2010semi,garg2022domain,wald2023birds}.
Mainstream PU learning methods \cite{du2014analysis,elkan2008learning,kiryo2017positive,zhao2022dist} deal with the absence of negative labels through risk estimator design, treating unlabeled samples as negatives and labeled samples as weighted combinations of positives and negatives. 
These risk estimators assume the class priors are given.
In practice, the class priors have to be estimated from data by mixture proportion estimation (MPE) techniques \cite{garg2021mixture,yao2022rethinking}.
These PU learning approaches are based on the Selected Completely at Random (SCAR) assumption \cite{elkan2008learning}.
In our context, SCAR would assume that every node from a non-novel category has an equal probability of being in the source domain. 

\subsection{Subpopulation Shift and PU Learning}
Subpopulation shift \cite{koh2021wilds,santurkar2021breeds,yang2023change} is a specific type of distribution shift where the proportions of categories differ between the source and target domains.
Prior work in subpopulation shift mostly focuses on learning a classifier with a decent worst group accuracy in the target domain \cite{liu2021just,Sagawa*2020Distributionally,yao2022improving}.
Our focus is different as we are interested in detecting the emergence of new categories in the face of subpopulation shifts among the pre-existing ones.

SCAR would break when there exists a subpopulation shift between the source and target. 
Using Fig. \ref{fig:problem} as an example, a kitchen product would have a higher probability of being in the source domain than a sports product due to the subpopulation shift, violating SCAR.
We show this connection more formally in Appendix \ref{app:scar_break}.
One line of work \cite{bekker2019beyond,gerych2022recovering,wang2024pue} relaxes SCAR through learning the propensity scores, i.e. the probability of a positive sample being labeled, and incorporating them into the risk estimator.
However, they require other assumptions in order to estimate propensity scores.
A relatively mild assumption is that the propensity score depends on fewer attributes than the PU classifier \cite{bekker2019beyond} while some stronger ones are Local Certainty and Probabilistic Gap \cite{gerych2022recovering,wang2024pue}.
One other recent work \cite{wald2023birds} uses a constrained learning method, CoNoC, to minimize the error of classifying labeled positive data as negative while reserving enough unlabeled data as negative.
The constrained learning framework has PAC-like finite sample guarantees but its empirical success has only been shown on tabular data so far.
Our method, RECO-SLIP, leverages the constrained learning framework due to its suitability to our problem.
In addition, RECO-SLIP uses a selective link prediction strategy to preserve the novel category subgroup structure in the node representation space for better separation.

\subsection{PU Learning on Graphs}
Graph PU learning utilizes the edge relation between node samples to learn a PU classifier.
LSDAN \cite{wu2019long} is the first method proposed for PU learning on attributed graphs.
LSDAN trains a long-short distance attention model with the non-negative PU loss \cite{kiryo2017positive}.
LP-PUL \cite{carnevali2021graph} uses the average shortest path distances from the positive nodes to identify the most negative nodes and performs label propagation to obtain the final predictions.
GRAB \cite{yoo2021accurate} estimates the class prior and learns the PU classifier through iterative belief propagation on the graph.
PU-GNN \cite{yang2023positive} is a state-of-the-art graph PU learning method.
PU-GNN segregates unlabeled nodes into two sets by proximity to source nodes and aligns the expectations of predicted label distributions with the class priors separately on the two sets.
It also employs structural regularization, which is similar to a link prediction loss.
However, its edge sampling space is all potential edges, making the regularization sample inefficient.
In contrast, RECO-SLIP reduces the edge sampling space for link prediction by identifying the subgraph that the classifier would underperform and needs further preservation of the subgroup structure.

\subsection{Anomaly Detection and OOD Detection on Graphs}
We briefly clarify the differences between our problem and node-level anomaly detection and out-of-distribution (OOD) detection.
Node-level anomaly detection \cite{ding2019deep,li2019specae,ma2021comprehensive} aims to identify a set of anomalous nodes or rank nodes in a given graph according to the degree of abnormality.
There are two major differences between anomaly detection and our problem.
Firstly, the concept of anomaly is at the level of individual nodes whereas the concept of novelty in our problem is at the level of node categories.
Secondly, there is no notion of source and target in anomaly detection.
This difference prevents node-level anomaly detection algorithms from leveraging domain labels and utilizing PU learning techniques.

Node-level OOD detection \cite{stadler2021graph,wu2023energybased,zhao2020uncertainty} is the task of detecting if a node is outside of the training distribution.
Most approaches leverage the uncertainty estimates from the probabilistic predictions of a graph neural network to quantify if a node is OOD.
There are two key differences between our problem and OOD detection.
Firstly, out-of-distribution nodes are not seen during training for OOD detection.
In our problem, novel categories are present during training in the target domain but the learning algorithm does not know which target nodes belong to the novel category.
Secondly, in-distribution category labels are typically provided in OOD detection whereas they are not provided in our problem setting.
The only type of labels we have is the source/target domain labels.

\section{Problem Formulation}
We consider a set of nodes $\mathcal{V}$ that can be partitioned into a subset of source nodes $\mathcal{V}_{S}$ and another subset of target nodes $\mathcal{V}_{T}$, i.e. $\mathcal{V}=\mathcal{V}_{S} \cup \mathcal{V}_{T}, \mathcal{V}_{S} \cap \mathcal{V}_{T} = \varnothing$.
Each node $v$ is accompanied by a feature vector $\mathbf{x}_v$.
The source nodes are generated from a source distribution $P_S$, which is a mixture of $K$ category distributions $G_1, G_2, \ldots, G_K$.
The target nodes consist of nodes from the aforementioned categories and those from a novel category that does not appear in the source.
We denote the target distribution by $P_T$.
Among the target distribution, we denote the distribution of the non-novel categories by $P_{T,0}$ and the distribution of the novel category by $P_{T,1}$.
Let $\alpha$ be the novel ratio in the target, i.e. $P_T = (1 - \alpha) P_{T,0} + \alpha P_{T,1}$.
We consider the scenario where subpopulation shifts may happen between the source and target among the non-novel categories.
More concretely, let $\boldsymbol{\gamma}, \hat{\boldsymbol{\gamma}} \in \Delta^{K-1}$ be probability vectors that determine the mixture proportions of the non-novel categories in source and target, respectively. Then,
\begin{equation}
    P_S=\sum_{i=1}^{K} \gamma_{i} G_{i}, P_{T,0}=\sum_{i=1}^{K} \hat{\gamma}_{i} G_{i}, \boldsymbol{\gamma} \neq \hat{\boldsymbol{\gamma}}
\end{equation}

Apart from the nodes, we also have access to an edge set $\mathcal{E} \subseteq \mathcal{V} \times \mathcal{V}$.
We assume the graph to be homophilous, meaning that two nodes having an edge between them have a higher probability of belonging to the same category than those that do not.
This assumption is common for large-scale graphs such as online social networks, citation graphs, or co-purchasing networks.


The goal of a learning algorithm for novel node category detection is to take the graph $\mathcal{G}=(\mathcal{V}_S, \mathcal{V}_T, \mathcal{E})$ and node features $\{\mathbf{x}_v \mid v \in \mathcal{V}\}$ as input, and learn a binary classifier $f: \mathcal{V} \rightarrow [0,1]$ that minimizes the following expected risk over the target distribution:
\begin{equation}
    R_{T}(f) = (1 - \alpha) \mathbb{E}_{v \sim P_{T,0}}[f(v)] + \alpha \mathbb{E}_{v \sim P_{T,1}}[1 - f(v)]
\end{equation}
Simply put, it would be ideal for the classifier to output a score close to $0$ if an input target node is from a non-novel category and output a score close to $1$ otherwise.
It is worth noting that the learning algorithm has access to the domain labels, i.e. whether a node belongs to $\mathcal{V}_S$ or $\mathcal{V}_T$, but not the category labels.
To jointly utilize the graph structure and node features, we mainly consider binary classifiers composed of a graph neural network (GNN) encoder $g$ and a multi-layer perceptron (MLP) head $h$, i.e. $f = h \circ g$.
Although the concept of novel/non-novel is at the category level, we will also say a node is novel/non-novel if it is from a novel/non-novel category for brevity in the remaining sections.

\section{Recall-Constrained Optimization with Selective Link Prediction (RECO-SLIP)}
\subsection{Recall-Constrained Optimization}
Recently, Wald et al. \cite{wald2023birds} propose a constrained optimization-based method as a principled alternative to PU learning under distribution shift.
They show that the expected target risk can be bounded by the false positive rate (FPR) \footnote{The positive in false positive rate refers to the novel category being detected, not the positive in PU-learning. We use FPR in the remaining paper to avoid confusion.} on the source domain, the negative of the recall on the target domain, and the divergence between the source and target distributions.
Motivated by the bound, they suggest minimizing the FPR while keeping the recall above a certain value.
Due to the suitability of this recall-constrained method to our problem, we adopt the same principle and describe it more formally below.

Let $\beta(f) = \mathbb{E}_{v\sim P_{S}}[f(v)]$ be the FPR on the source domain, namely the error rate of identifying source nodes as novel.
Let $\alpha(f) = \mathbb{E}_{v\sim P_{T}}[f(v)]$ be the recall on the target domain, which is the rate of identifying target nodes as novel.
We denote the empirical estimate of $\alpha(f)$ and $\beta(f)$ by $\hat{\alpha}(f)$ and $\hat{\beta}(f)$.
Note that the FPR defined here is not evaluated on all non-novel nodes but only the non-novel nodes within the source domain.
Also, the recall here is not just evaluated on the novel nodes but all target nodes.
We do not have ground-truth novel/non-novel labels in the target domain so we could only use domain labels as a proxy.


We use the average output score of $f$ over the source nodes and the target nodes as a differentiable proxy of $\hat{\beta}(f)$ and $\hat{\alpha}(f)$, respectively.
Then, the optimization problem of minimizing the empirical FPR and keeping the empirical recall above a value $\Tilde{\alpha}$ can be written as:

\begin{align}\label{eq:opt_prob}
    &\min_{f} \frac{1}{|\mathcal{V}_S|} \sum_{v \in \mathcal{V}_S} f(v),
    \ 
    \text{s.t. } \frac{1}{|\mathcal{V}_T|} \sum_{v \in \mathcal{V}_T} f(v) \geq \Tilde{\alpha} 
\end{align}

\begin{figure}[t]
    \centering
    \includegraphics[width=0.9\textwidth]{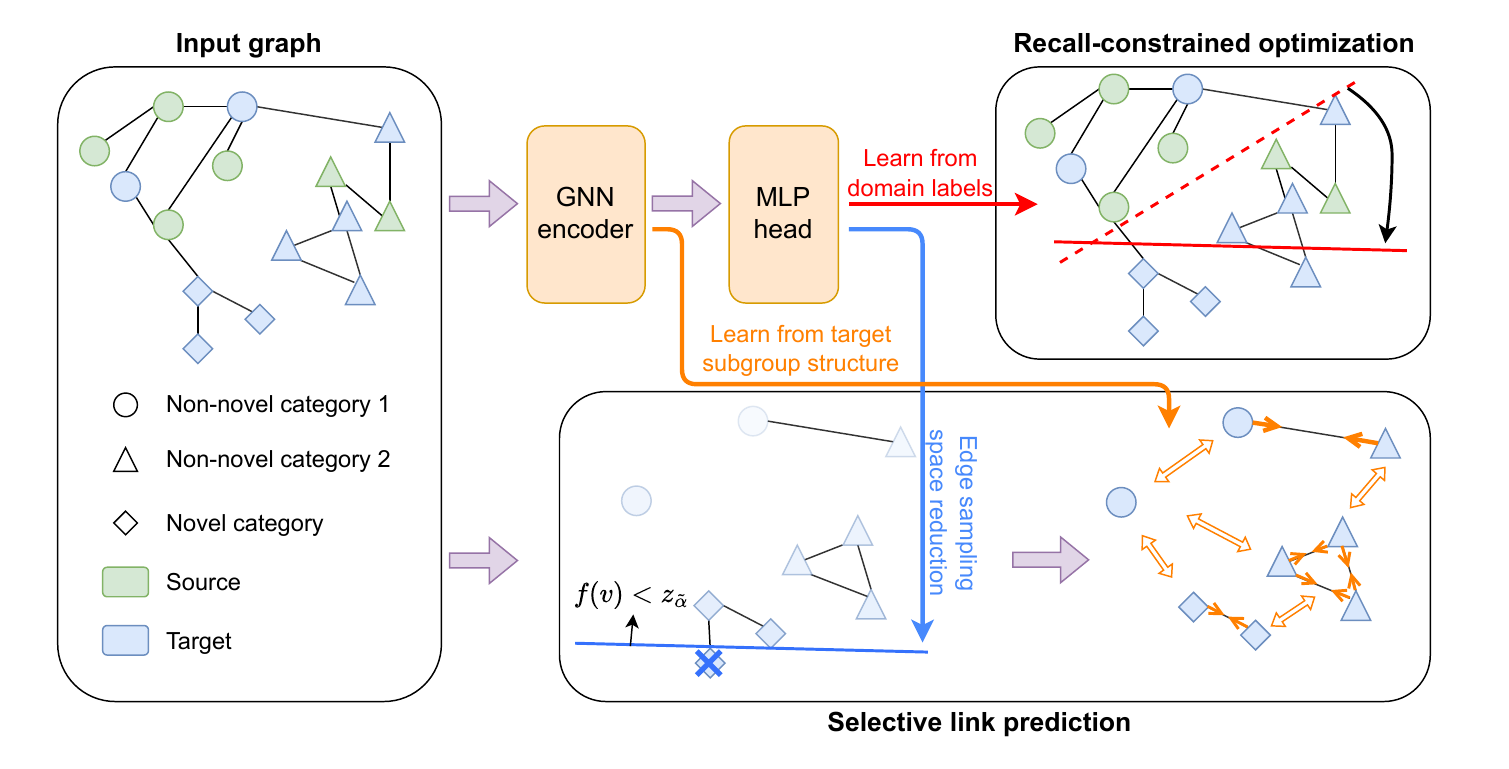}
    \caption{
    An illustration of RECO-SLIP. 
    The upper-right module is the recall-constrained optimization component (Eq. \ref{eq:opt_prob}) where the classifier adjusts its scores to minimize the FPR on the source while reserving enough target nodes as novel.
    The bottom module is the selective link prediction component where the link prediction loss (Eq. \ref{eq:lp}) is imposed on the target subgraph excluding the nodes with the highest scores (Eq. \ref{eq:z_alpha}, \ref{eq:e-}, \ref{eq:e+}).
    A solid orange arrow pair between two nodes denotes their representation similarity is maximized whereas a bidirectional hollow arrow denotes the similarity is minimized.
    }
    \label{fig:method}
\end{figure}

In practice, we solve the optimization problem using different values of $\Tilde{\alpha}$. 
Then we select the model that has the highest empirical recall out of the models that achieve an empirical FPR below a user-specified threshold $\Tilde{\beta}$.

The intuitive idea of why solving the constrained learning problem could outperform standard PU learning methods under subpopulation shift can be shown in the upper-right module of Fig. \ref{fig:method}.
Standard PU learning directly discriminates source from target nodes.
Since there are more target nodes in non-novel category 2, standard PU learning would result in the tilted decision boundary represented by the dashed line.
By solving the constrained learning problem, the classifier avoids identifying source nodes as novel while reserving enough target nodes as novel, resulting in a more horizontal decision boundary represented by the solid line.

\subsection{Selective Link Prediction}

Using the domain labels as a proxy of novel/non-novel labels essentially views all target nodes as equally novel.
The novel and non-novel node representations within the target domain will not be sufficiently separated since the domain labels confuse them as the same.
However, the edge connection pattern can reveal additional information due to the homophily property.
For instance, the target novel nodes should have few edge connections to the target non-novel nodes since they belong to different categories.
Therefore, we use link prediction with the graph autoencoder (GAE) \cite{kipf2016variational} objective as an auxiliary task during training to preserve the category subgroup structure in the node representation space.

Link prediction maximizes the similarity between two node representations if the two nodes are connected and minimizes their similarity otherwise.
To determine which pairs of unconnected nodes should have minimized representation similarities, we perform negative sampling from non-existent edges.
As real-world graphs are sparse, the negative sampling space is huge compared to the existent edges. 
Therefore, we reduce the sampling space according to our problem to improve sample efficiency.
The non-existent edges $\mathcal{E}^{\mathsf{C}}$ can be categorized into three subsets $\mathcal{E}^{\mathsf{C}}_{S, S}$, $\mathcal{E}^{\mathsf{C}}_{S, T}$, and $\mathcal{E}^{\mathsf{C}}_{T, T}$ defined as follows:

\begin{align*}
    &\mathcal{E}^{\mathsf{C}}_{S, S} = \{(v_i, v_j) \mid v_i \in \mathcal{V}_S \land v_j \in \mathcal{V}_S \land (v_i, v_j) \in \mathcal{E}^{\mathsf{C}}\} \\
    &\mathcal{E}^{\mathsf{C}}_{S, T} = \{(v_i, v_j) \mid ((v_i \in \mathcal{V}_S \land v_j \in \mathcal{V}_T) \lor (v_i \in \mathcal{V}_T \land v_j \in \mathcal{V}_S)) \land (v_i, v_j) \in \mathcal{E}^{\mathsf{C}}\} \\
    &\mathcal{E}^{\mathsf{C}}_{T, T} = \{(v_i, v_j) \mid v_i \in \mathcal{V}_T \land v_j \in \mathcal{V}_T \land (v_i, v_j) \in \mathcal{E}^{\mathsf{C}}\}
\end{align*}

\begin{algorithm}[t]
\caption{RECO-SLIP}\label{alg:reco_slip}
\begin{algorithmic}
\Require Dataset: $\mathcal{V}_S, \mathcal{V}_T, \mathcal{E}, \{\mathbf{x}_v \mid v \in \mathcal{V}_S \cup \mathcal{V}_T\}$, hyper-parameters: $\xi, \Tilde{\boldsymbol{\alpha}}, \Tilde{\beta}, T$
\For{$\Tilde{\alpha} \in \Tilde{\boldsymbol{\alpha}}$}
    \State Initialize binary classifier $f^{(0)}$ and dual variable $\lambda^{(0)}$.
    \For{$t \gets 1$ to $T$}
        \State Construct $\mathcal{E}_{-}$ and $\mathcal{E}_{+}$ based on Eq. \ref{eq:e-} and Eq. \ref{eq:e+}.
        \State Sample $\mathcal{E}_{-}^{*}$ from $\mathcal{E}_{-}$.
        \State $f^{(t)}, \lambda^{(t)} \gets$  Primal-dual optimization update with the Lagrangian in Eq. \ref{eq:lagrangian}.
    \EndFor
    \State $f_{\Tilde{\alpha}} \gets f^{(T)}$
    \State Calculate empirical recall $\hat{\alpha}(f_{\Tilde{\alpha}})$ and empirical FPR $\hat{\beta}(f_{\Tilde{\alpha}})$.
\EndFor
\State Return $\argmax_{f_{\Tilde{\alpha}}: \Tilde{\alpha} \in \Tilde{\boldsymbol{\alpha}}, \hat{\beta}(f_{\Tilde{\alpha}}) < \Tilde{\beta}} \hat{\alpha}(f_{\Tilde{\alpha}})$
\end{algorithmic}
\end{algorithm}

$\mathcal{E}^{\mathsf{C}}_{S, S}$ are non-existent edges where the nodes on both sides are from the source domain.
We do not sample from this subset because all source nodes are non-novel and there is no need to separate them apart.
$\mathcal{E}^{\mathsf{C}}_{S, T}$ are non-existent edges where the node on one side is from the source domain and the other is from the target domain.
We do not sample from this subset either because the main task using the domain labels is already separating source and target nodes.
The strategy above restricts us to sampling from the target subgraph.
We can further reduce the sampling space based on the score produced by the classifier.
When solving the constrained optimization in Eq. \ref{eq:opt_prob}, the classifier reserves at least $\Tilde{\alpha}$ portion of the target nodes as novel.
The other target nodes scored at the bottom $1 - \Tilde{\alpha}$ portion are the samples that the classifier is less confident in identifying as novel and require the auxiliary task to separate node representations of different subgroups.
Therefore, we reduce the negative sampling space to the non-existent edges among the target nodes scored at the bottom $1 - \Tilde{\alpha}$.
Let $z_{\Tilde{\alpha}}$ denote the score separating the top $\alpha$ and bottom $1 - \alpha$ target nodes, and $\mathcal{E}_{-}$ denote the final negative sampling space:
\begin{align}
    &z_{\Tilde{\alpha}} = \min \{ z \mid \frac{1}{|\mathcal{V}_{T}|} \sum_{v \in \mathcal{V}_T} \mathbbm{1}[f(v) \leq z] > 1 - \Tilde{\alpha}\} \label{eq:z_alpha} \\
    &\mathcal{E}_{-} = \{(v_i, v_j) \mid f(v_i) < z_{\Tilde{\alpha}} \land f(v_j) < z_{\Tilde{\alpha}} \land (v_i, v_j) \in \mathcal{E}^{\mathsf{C}}_{T, T} \} \label{eq:e-}
\end{align}
To produce a balancing force towards the separation induced by $\mathcal{E}_{-}$, we consider the existing edges between the same nodes involved in $\mathcal{E}_{-}$ the positive samples:
\begin{align}\label{eq:e+}
    &\mathcal{E}_{+} = \{(v_i, v_j) \mid f(v_i) < z_{\Tilde{\alpha}} \land f(v_j) < z_{\Tilde{\alpha}} \land v_i \in \mathcal{V}_T \land v_j \in \mathcal{V}_T \land (v_i, v_j) \in \mathcal{E} \}
\end{align}

At each iteration, we uniformly sample the same amount of negative edges as the positive samples from $\mathcal{E}_{-}$.
We denote the sampled negative edges by $\mathcal{E}_{-}^{*}$.
Let $\mathbf{g}_{v} = g(v)$ denote the representation of node $v$ produced by the GNN encoder and $\sigma$ denote the sigmoid function.
The final link prediction loss is shown below:
\begin{align}\label{eq:lp}
    \mathcal{L}_{lp}(g) =  - \frac{1}{|\mathcal{E}_{+}|} \sum_{(v_i, v_j) \in \mathcal{E}_{+}} \log \sigma(\mathbf{g}_{v_i} \cdot \mathbf{g}_{v_j}) - \frac{1}{|\mathcal{E}_{-}^{*}|} \sum_{(v_i, v_j) \in \mathcal{E}_{-}^{*}} \log(1 - \sigma(\mathbf{g}_{v_i} \cdot \mathbf{g}_{v_j}))
\end{align}

We multiply the link prediction loss with a hyper-parameter $\xi$ and add it to the constrained learning objective.
Then the Lagrangian can be computed as follows:
\begin{align}\label{eq:lagrangian}
    \mathcal{L}(g, h, \lambda) = \hat{\beta}(h \circ g) + \xi \mathcal{L}_{lp}(g) + \lambda (\Tilde{\alpha} - \hat{\alpha}(h \circ g))
\end{align}
where $\lambda$ is a non-negative dual variable.
We apply $T$ primal-dual optimization \cite{gallegoPosada2022cooper} updates to learn the classifier.
The overall procedure is in Algorithm \ref{alg:reco_slip}.

\section{Experiments}

\subsection{Experimental Setup}

\begin{table}[t]
\caption{Source ratio per category for Cora-S, CiteSeer-S, Computers-S, and Photo-S. Each entry represents the ratio of nodes belonging to a category that is assigned to the source domain. The source ratio of the last available category in each dataset is $0$ because the category is novel.}
\label{tab:src_ratio}
\begin{center}
\addtolength{\tabcolsep}{2pt}  
\scalebox{0.95}{
\begin{tabular}{lcccccccccc}
\toprule
& \multicolumn{10}{c}{\textbf{Category label}} \\
\midrule
                                   & 1          & 2          & 3          & 4          & 5          & 6          & 7    & 8    & 9    & 10 \\
\midrule
Cora-S                             & $0.1$      & $0.9$      & $0.1$      & $0.9$      & $0.1$      & $0.9$      & $0$   & -     & -    & -  \\
CiteSeer-S                         & $0.9$      & $0.1$      & $0.9$      & $0.1$      & $0.5$      & $0$        & -     & -     & -    & -  \\
Computers-S                        & $0.1$      & $0.9$      & $0.1$      & $0.9$      & $0.1$      & $0.9$      & $0.1$ & $0.9$ & $0.5$& $0$ \\
Photo-S                            & $0.9$      & $0.1$      & $0.9$      & $0.1$     & $0.9$      & $0.1$      & $0.5$ & $0$ & -    & - \\
\bottomrule
\end{tabular}}
\addtolength{\tabcolsep}{-2pt}
\end{center}
\end{table}

\begin{table}[t]
\caption{Dataset statistics.}
\label{data_stat}
\begin{center}
\addtolength{\tabcolsep}{1pt}
\scalebox{0.9}{
\begin{tabular}{lccccc}
\toprule
               &\multicolumn{1}{c}{\bf Cora-S}  &\multicolumn{1}{c}{\bf Citeseer-S}  &\multicolumn{1}{c}{\bf Computers-S}  &\multicolumn{1}{c}{\bf Photo-S} &\multicolumn{1}{c}{\bf arxiv}
\\ \midrule 
\# categories                        & 7                            & 6                                & 10                                & 8        & 6 \\
\# source nodes                      & 1317                         & 1265                             & 5252                              & 3068     & 859 \\
\# target nodes                      & 1391                         & 2062                             & 8500                              & 4582     & 3442 \\
\# novel nodes                       & 180                          & 508                              & 291                               & 331      & 176  \\
Novel ratio (\# novel / \# target)   & 0.129                        & 0.246                            & 0.034                             & 0.072    & 0.051  \\
\bottomrule
\end{tabular}}
\addtolength{\tabcolsep}{-1pt}
\end{center}
\end{table}

\subsubsection{Data}~
We evaluate RECO-SLIP and baseline methods on five widely used public benchmark datasets: Cora \cite{sen2008collective}, CiteSeer \cite{sen2008collective}, Computers \cite{shchur2018pitfalls}, Photo \cite{shchur2018pitfalls}, and arxiv \cite{hu2020open}.
Cora, CiteSeer, and arxiv are academic citation graphs while Computers and Photo are product co-purchasing networks.
We use the original labels provided by the datasets as the category labels.
For Cora, CiteSeer, Computers, and Photo, we view the last class as the novel category and simulate subpopulation shifts by splitting nodes in each category into source and target based on Table \ref{tab:src_ratio}.
We name the preprocessed versions of the four datasets with ``-S'' at the end representing ``shift''.
arxiv is a subset of the ogbn-arxiv graph \cite{hu2020open} containing $6$ robotics-related categories from $1990$ to $2012$.
arxiv enables a natural source/target split and a novel category emergence since each node is associated with a timestamp and nodes belonging to the category ``cs.SY: systems and control'' did not exist until $2007$.
Therefore, we define nodes with timestamps before $2007$ as the source and the others as the target.
In terms of train/test data split, we consider the commonly used transductive node classification setup. 
The test nodes of arxiv are the nodes with timestamps in $2012$ while the test nodes of the other four datasets are randomly selected from the target domain.
We further split a part of the training nodes into the validation set for model selection.
We show the dataset-specific statistics in Table \ref{data_stat}.

\subsubsection{Baselines}~
We compare RECO-SLIP with PU learning approaches both SCAR and not SCAR-based as well as graph PU learning approaches.
\begin{itemize}
    \item \textbf{Domain discriminator}: The domain discriminator directly discriminates source from target nodes. It is the basis of most PU-learning approaches.
    \item \textbf{uPU} \cite{du2014analysis}: uPU is an unbiased risk estimator for PU-learning. It treats unlabeled samples as negative and labeled samples as weighted combinations of positives and negatives.
    \item \textbf{nnPU} \cite{kiryo2017positive}: nnPU is a non-negative risk estimator for PU-learning. It addresses the overfitting problem of uPU since the empirical risk of uPU could be negative and unbounded below. 
    \item \textbf{SAR-EM} \cite{bekker2019beyond}: SAR-EM is a propensity-weighting approach for PU-learning when SCAR does not hold. It jointly learns the PU classifier and the propensity score of each data point via the EM algorithm. We select this method as the representative method of propensity-weighting approaches due to its relatively flexible assumptions and publicly available code implementation. Later work using propensity-weighting \cite{gerych2022recovering,wang2024pue} imposes strong assumptions such as Local Certainty and Probabilistic Gap\footnote{Local Certainty assumes the relationship between the observed features and the true class is a deterministic function, meaning that the class distributions do not overlap. Probabilistic Gap allows overlapping class distributions but assumes that the propensity scores follow the ordering of the posterior probabilities, i.e. $p(y=1|\mathbf{x})$.} and do not provide code. Therefore, we omit these methods in our experiments.
    \item \textbf{LP-PUL} \cite{carnevali2021graph}: LP-PUL is a graph-based PU learning approach. It first identifies the furthermost target nodes from the source nodes as the most novel nodes based on the shortest path distances, then performs label propagation on top of the graph. Note that LP-PUL is the only experimented method that does not learn a graph neural network.
    \item \textbf{PU-GNN} \cite{yang2023positive}: PU-GNN is a state-of-the-art graph PU learning method. It adapts the Dist-PU risk estimator \cite{zhao2022dist} towards graph data and applies structural regularization to learn pairwise relations between nodes.
\end{itemize}
Aside from the methods mentioned above, we also train a classifier using the Oracle novel/non-novel labels as an upper bound reference.

\subsubsection{Hyper-Parameters}~
All methods except LP-PUL use a two-layer graph convolution network (GCN) \cite{kipf2017semisupervised} coupled with a two-layer MLP as the classifier architecture.
We find that the Adam optimizer \cite{kingma2014adam} with a learning rate of $0.001$ works well with the domain discriminator components in all methods except LP-PUL.
This optimizer setting also works well for the primal and dual optimizers of RECO-SLIP.
For hyper-parameters specific to RECO-SLIP, we set $\xi = 0.001$, $T = 1000$, and $\Tilde{\boldsymbol{\alpha}}=[0.05, 0.1, 0.15, 0.2, 0.25]$ for all datasets.
By default, we set $\Tilde{\beta}$ to $0.01$.
However, since the empirical FPR estimate under all $\Tilde{\alpha}$ could not go below $0.01$ on Photo-S and CiteSeer-S due to dataset characteristics, we set $\Tilde{\beta}$ to $0.05$ for these two datasets.

\subsubsection{Evaluation Metrics}~
Following the convention of novelty and anomaly detection, we use AU-ROC to evaluate the ability of a classifier to rank novel nodes above non-novel nodes.
We test all methods with 10 different random seeds and report the average scores and standard errors.

We list additional details of the experimental setup in Appendix \ref{app:dataset} and \ref{app:hyperparam}.

\subsection{Results and Discussions}

\begin{table}[t]
\caption{AU-ROC on five datasets. The best performance on each dataset is highlighted in \textbf{bold} and the second best performance is highlighted by \underline{underline}.}
\label{auroc-table}
\begin{center}
\addtolength{\tabcolsep}{1pt}
\scalebox{0.9}{
\begin{tabular}{lccccc}
\toprule
\multicolumn{1}{c}{\bf Method}  &\multicolumn{1}{c}{\bf Cora-S}  &\multicolumn{1}{c}{\bf CiteSeer-S}  &\multicolumn{1}{c}{\bf Computers-S}  &\multicolumn{1}{c}{\bf Photo-S} &\multicolumn{1}{c}{\bf arxiv}
\\ \midrule 
Oracle                           & 0.953{\scriptsize$\pm$0.004}& 0.864{\scriptsize$\pm$0.011}     & 0.986{\scriptsize$\pm$0.005}      & 0.985{\scriptsize$\pm$0.001}    & 0.847{\scriptsize$\pm$0.011} \\
\midrule
Domain discriminator             & 0.705{\scriptsize$\pm$0.016}& \underline{0.717}{\scriptsize$\pm$0.011}     & \underline{0.877}{\scriptsize$\pm$0.025}      & 0.671{\scriptsize$\pm$0.035}    & 0.628{\scriptsize$\pm$0.020}\\
uPU                              & 0.705{\scriptsize$\pm$0.016}& 0.701{\scriptsize$\pm$0.023}     & 0.834{\scriptsize$\pm$0.032}      & 0.677{\scriptsize$\pm$0.038}    & 0.628{\scriptsize$\pm$0.018}\\
nnPU                             & 0.705{\scriptsize$\pm$0.015}& 0.701{\scriptsize$\pm$0.023}     & 0.830{\scriptsize$\pm$0.031}      & 0.676{\scriptsize$\pm$0.037}    & 0.629{\scriptsize$\pm$0.020}\\
SAR-EM                           & \underline{0.740}{\scriptsize$\pm$0.039}& 0.695{\scriptsize$\pm$0.015}     & 0.721{\scriptsize$\pm$0.073}      & 0.607{\scriptsize$\pm$0.036}    & 0.615{\scriptsize$\pm$0.032}\\
LP-PUL                           & 0.666{\scriptsize$\pm$0.000}& 0.638{\scriptsize$\pm$0.000}     & 0.830{\scriptsize$\pm$0.000}      & 0.256{\scriptsize$\pm$0.000}    & \underline{0.658}{\scriptsize$\pm$0.000}\\
PU-GNN                           & 0.705{\scriptsize$\pm$0.015}& 0.701{\scriptsize$\pm$0.024}     & 0.833{\scriptsize$\pm$0.034}      & \underline{0.678}{\scriptsize$\pm$0.040}    & 0.627{\scriptsize$\pm$0.018}
\\ \midrule
RECO-SLIP                        & \textbf{0.770}{\scriptsize$\pm$0.025}& \textbf{0.745}{\scriptsize$\pm$0.019}& \textbf{0.948}{\scriptsize$\pm$0.008}& \textbf{0.810}{\scriptsize$\pm$0.011}   & \textbf{0.710}{\scriptsize$\pm$0.022}\\
\bottomrule
\end{tabular}}
\addtolength{\tabcolsep}{-1pt}
\end{center}
\end{table}

The overall results are shown in Table \ref{auroc-table}.
One can have an estimate of how good the domain labels are as a proxy of the novel/non-novel labels by examining the performance gap between Oracle and the domain discriminator.
For example, the performance gap on Photo-S is above $0.3$ while the ones on other datasets range from $0.1$ to $0.25$, indicating that Photo-S is harder for PU learning.
Out of all methods, RECO-SLIP is the best-performing one on all datasets, demonstrating its effectiveness in novel node category detection under subpopulation shifts.

Standard PU learning approaches such as the domain discriminator, uPU, and nnPU perform relatively stable across datasets.
Perhaps because AU-ROC is a ranking metric, the advantage of avoiding the predictions to be overly novel of uPU and nnPU is not shown.
In addition, uPU and nnPU require mixture proportion estimation in the warm-up phase, which could divert the PU classifier learning process and lead to a slight performance decrease from the domain discriminator.
RECO-SLIP outperforms these three methods since they do not address the subpopulation shift problem and utilize the subgroup information provided by the graph structure.

SAR-EM addresses subpopulation shifts by jointly learning the propensity scores and the PU classifier.
However, it is extremely unstable due to the application of the EM algorithm to neural networks.
Its instability is reflected in the results.
SAR-EM achieves the second highest AU-ROC on Cora-S but gets the worst AU-ROC on Computers-S and arxiv.
LP-PUL utilizes label propagation to encourage each subgroup to have consistent predictions.
Nevertheless, it uses the shortest path distances to identify the most novel nodes and does not leverage the node feature information.
This approach can be brittle when the node features are informative but the edge connections are noisy.
For instance, LP-PUL performs worse than a no-skill classifier on Photo-S, potentially because it identifies the incorrect novel nodes in the initialization phase.
PU-GNN leverages an adapted Dist-PU risk estimator coupled with structural regularization, which is similar to the idea of link prediction.
However, it does not outperform standard PU learning approaches very much because it is also based on SCAR and the sampling space of its structural regularization is all potential edges, leading to low sample efficiency.
RECO-SLIP simultaneously addresses subpopulation shifts and utilizes the subgroup information provided by the graph structure in a stable and sample-efficient manner, resulting in higher performance.

\begin{table}[t]
\caption{Ablation study on selective link prediction evaluated by AU-ROC. The best performance is highlighted in \textbf{bold} and the second best is highlighted by \underline{underline}.}
\label{tab:ablation-auroc}
\begin{center}
\scalebox{0.93}{
\begin{tabular}{lccccc}
\toprule
\multicolumn{1}{c}{\bf Method}  &\multicolumn{1}{c}{\bf Cora-S}  &\multicolumn{1}{c}{\bf CiteSeer-S}  &\multicolumn{1}{c}{\bf Computers-S}  &\multicolumn{1}{c}{\bf Photo-S}  &\multicolumn{1}{c}{\bf arxiv}
\\ \midrule 
w/o link prediction             & 0.755{\scriptsize$\pm$0.023}& 0.714{\scriptsize$\pm$0.035}     & \underline{0.946}{\scriptsize$\pm$0.006}      & 0.795{\scriptsize$\pm$0.013}      & \underline{0.703}{\scriptsize$\pm$0.019}\\
w/ full link prediction         & 0.750{\scriptsize$\pm$0.031}& 0.719{\scriptsize$\pm$0.030}     & 0.943{\scriptsize$\pm$0.013}      & 0.793{\scriptsize$\pm$0.012}      & 0.693{\scriptsize$\pm$0.024}\\
w/ target link prediction       & \underline{0.767}{\scriptsize$\pm$0.027}& \underline{0.723}{\scriptsize$\pm$0.032}     & 0.943{\scriptsize$\pm$0.011}      & \textbf{0.812}{\scriptsize$\pm$0.010}      & 0.699{\scriptsize$\pm$0.020}\\
RECO-SLIP                       & \textbf{0.770}{\scriptsize$\pm$0.025}& \textbf{0.745}{\scriptsize$\pm$0.019}& \textbf{0.948}{\scriptsize$\pm$0.008} & \underline{0.810}{\scriptsize$\pm$0.011}      & \textbf{0.710}{\scriptsize$\pm$0.022}\\
\bottomrule
\end{tabular}}
\end{center}
\end{table}

\subsection{Auxiliary Experiments}
\subsubsection{Ablation Study}~
We conduct an ablation study on the selective link prediction component of RECO-SLIP.
To understand the effectiveness of link prediction, we consider dropping the link prediction loss (w/o link prediction).
In addition, we consider vanilla full link prediction without reducing the edge sampling space (w/ full link prediction) and doing link prediction on the target subgraph without using the classifier score filtering rule shown in Eq. \ref{eq:z_alpha} and \ref{eq:e-} (w/ target link prediction).
We show the results in Table \ref{tab:ablation-auroc}.
RECO-SLIP performs the best across datasets except for slightly under-performing w/ target link prediction on Photo-S.
This is potentially because Photo-S is the hardest dataset for PU learning and relying on the classifier scores for filtering does not further improve from target subgraph sampling.
It is worth noting that w/ target link prediction also performs the second best on Cora-S and CiteSeer-S.
This demonstrates the effectiveness of not sampling source-source and source-target pairs.

\begin{figure*}[t]
        \centering
        \begin{subfigure}[b]{0.43\textwidth}
            \centering
            \includegraphics[width=\textwidth]{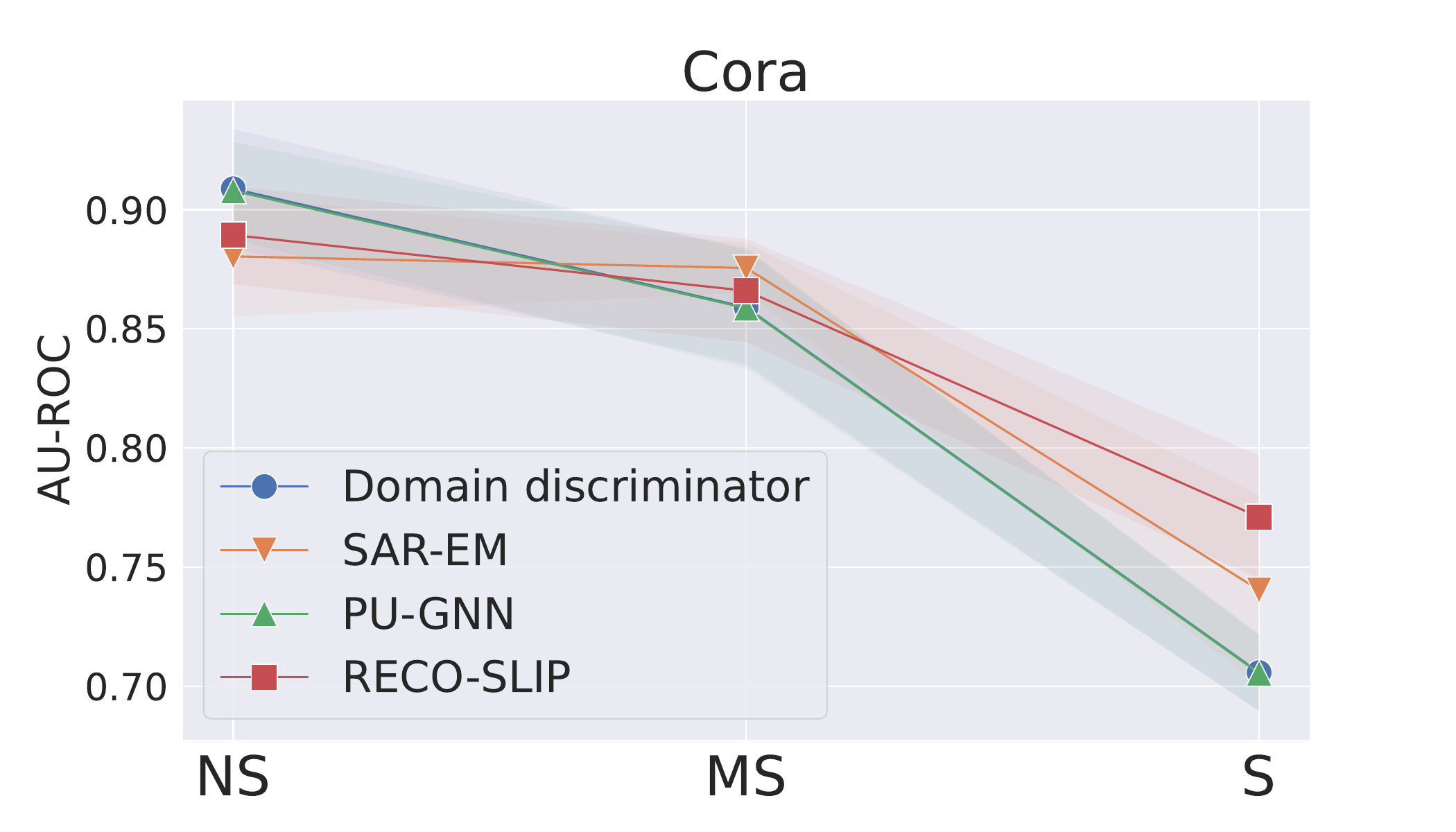}
            {{\small}}    
            \label{fig:cora_shift_intensity}
        \end{subfigure}
        \begin{subfigure}[b]{0.43\textwidth}  
            \centering 
            \includegraphics[width=\textwidth]{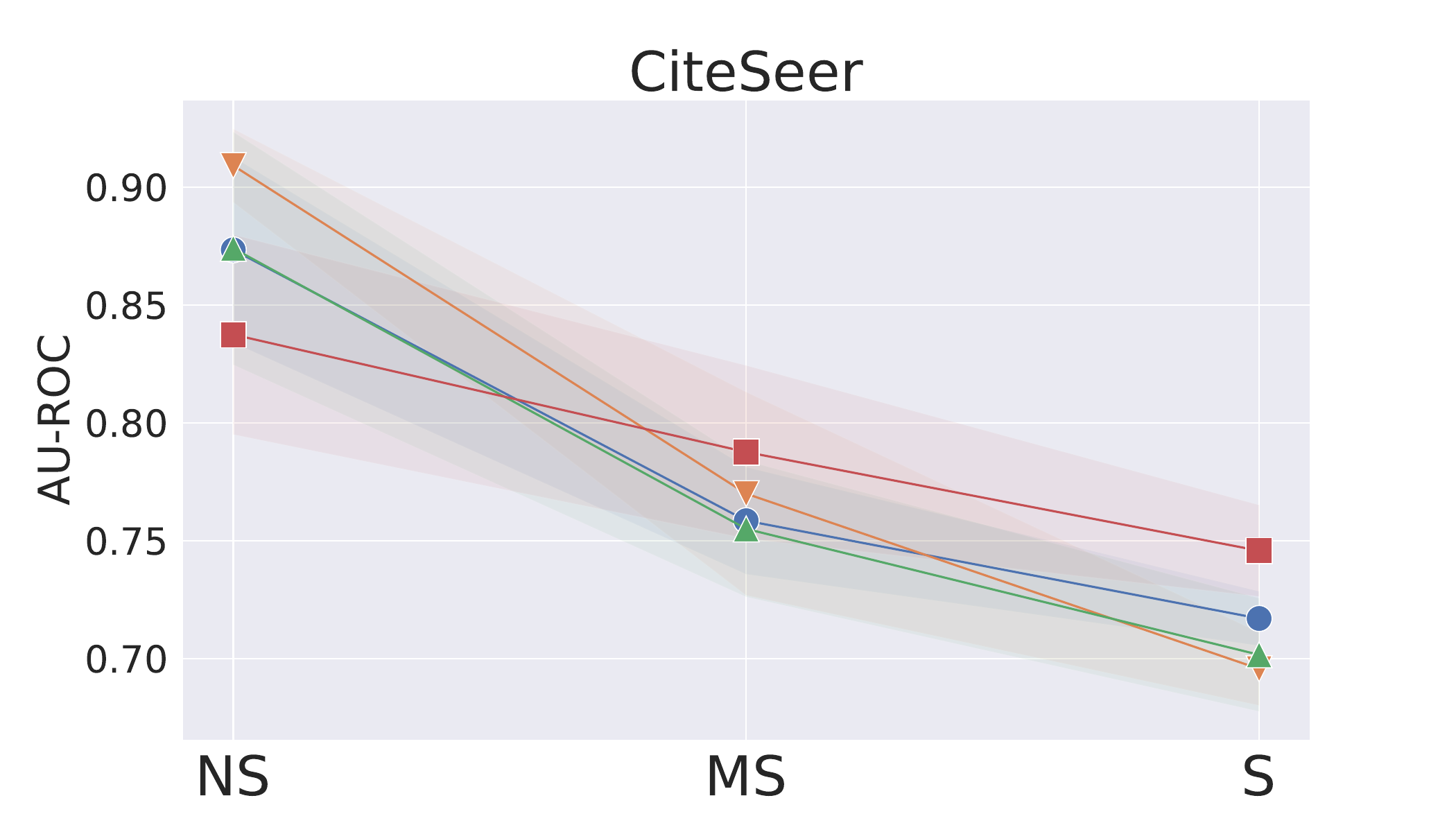}
            {{\small}}    
            \label{fig:citeseer_shift_intensity}
        \end{subfigure}
        \vskip \baselineskip
        \begin{subfigure}[b]{0.43\textwidth}
            \centering 
            \includegraphics[width=\textwidth]{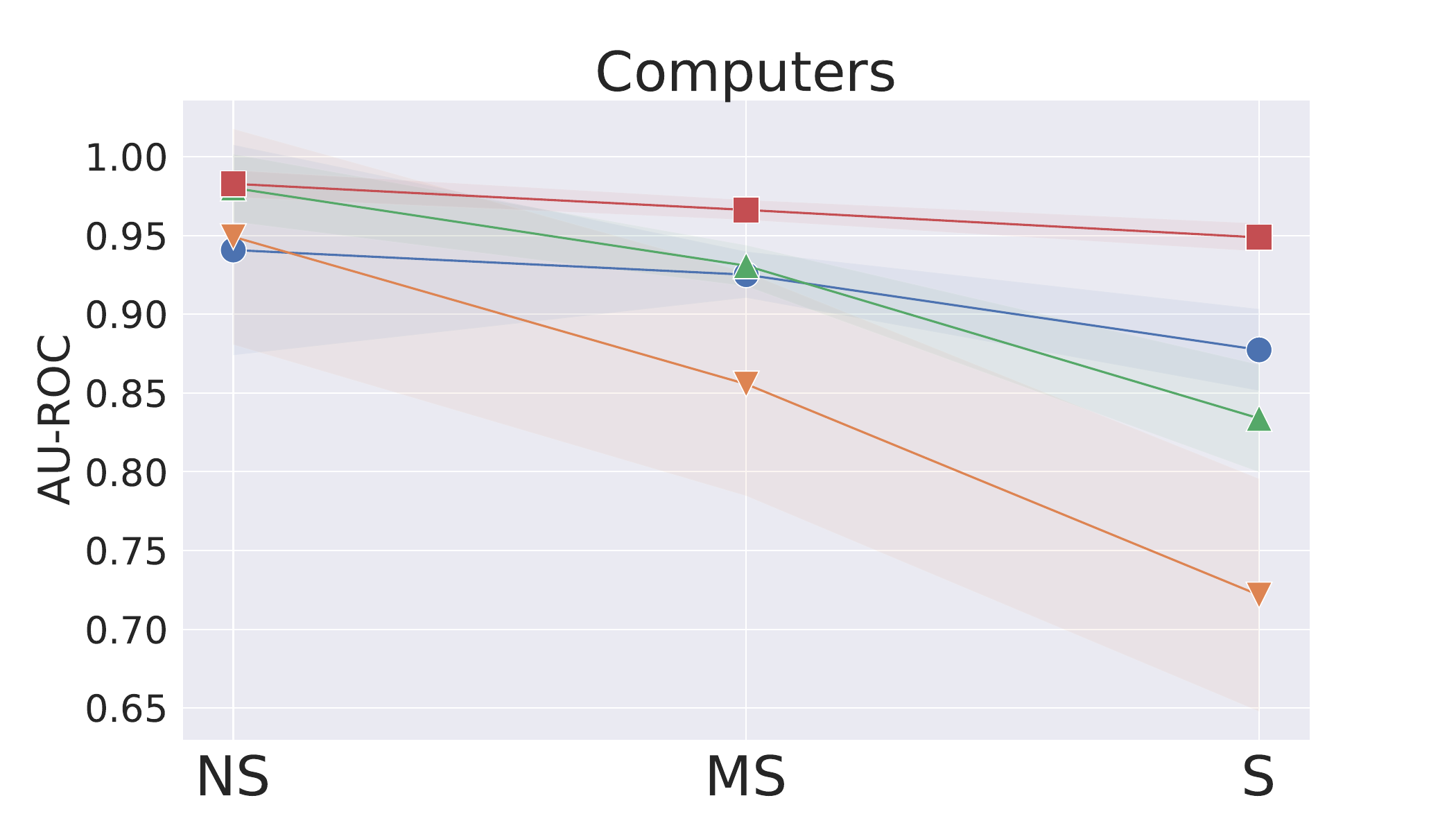}
            {{\small}}    
            \label{fig:computers_shift_intensity}
        \end{subfigure}
        \begin{subfigure}[b]{0.43\textwidth}   
            \centering 
            \includegraphics[width=\textwidth]{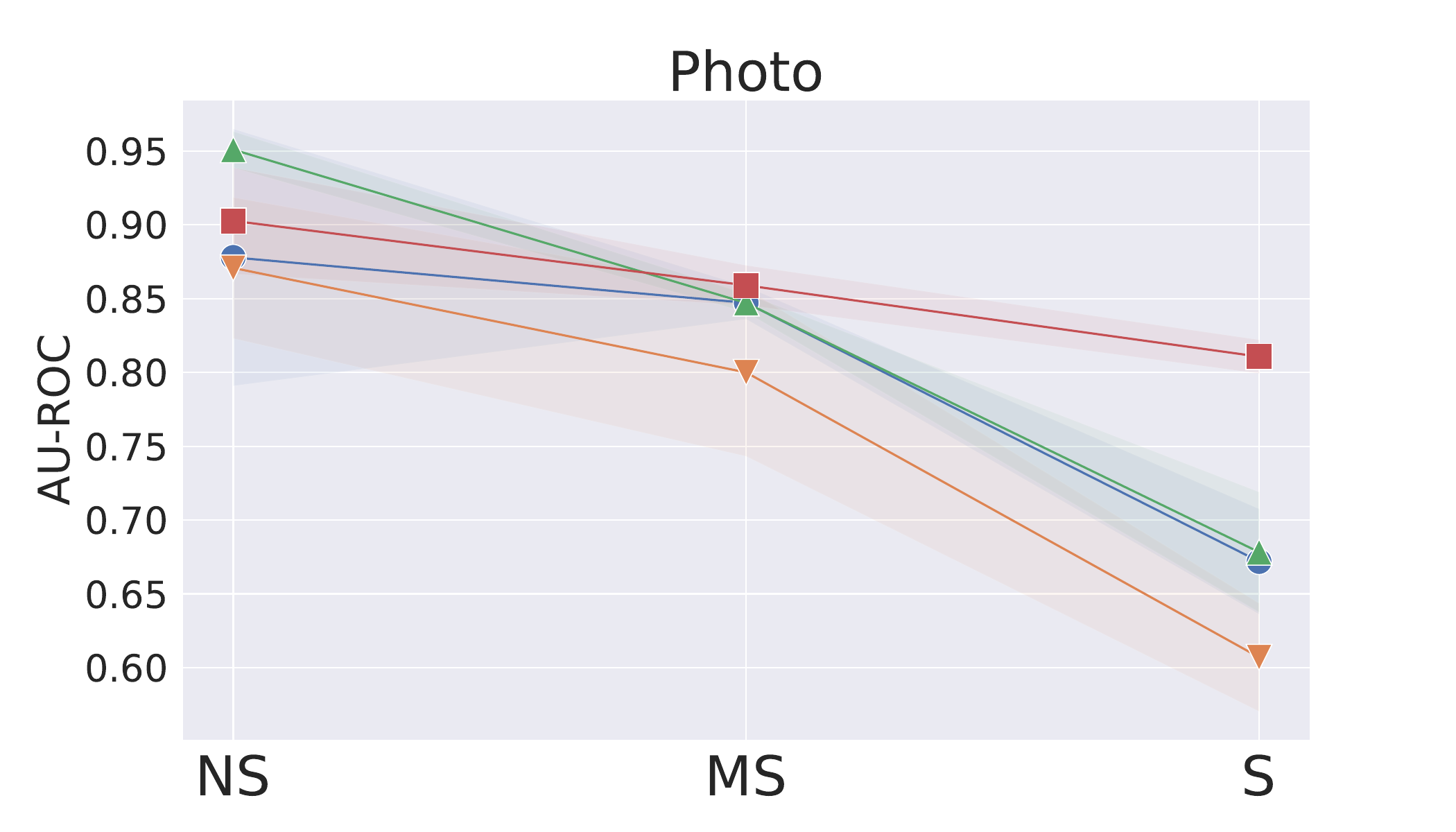}
            {{\small}}    
            \label{fig:photo_shift_intensity}
        \end{subfigure}
        \caption[]
        {\small Overall results of the shift intensity study. The x-axis represents shift intensity (NS: no shift, MS: minor shift, S: shift) and the y-axis represents AU-ROC performance.}
        \label{fig:shift_intensity}
\end{figure*}

\begin{table}[t]
\caption{Average performance rank of representative methods under different shift intensities (highest rank in \textbf{bold}, second-highest \underline{underlined}).}
\label{tab:intensity_rank}
\begin{center}
\addtolength{\tabcolsep}{3pt}
\scalebox{0.9}{
\begin{tabular}{lcccc}
\toprule
\multicolumn{1}{c}{\bf Method}  &\multicolumn{1}{c}{\bf NS}  &\multicolumn{1}{c}{\bf MS}  &\multicolumn{1}{c}{\bf S}  &\multicolumn{1}{c}{\bf Overall}
\\ \midrule 
Domain discriminator             & 2.75                      & \underline{2.75}           & \underline{2.50}          & \underline{2.66} \\
SAR-EM                           & 3.00                      & \underline{2.75}           & 3.50                      & 3.08 \\
PU-GNN                           & \textbf{1.75}             & 3.25                       & 3.00                      & \underline{2.66} \\
RECO-SLIP                        & \underline{2.50}          & \textbf{1.25}              & \textbf{1.00}             & \textbf{1.58} \\
\bottomrule
\end{tabular}}
\addtolength{\tabcolsep}{-3pt}
\end{center}
\end{table}

\subsubsection{Shift Intensity Study}~
We further study how different classes of methods react to the shift intensity on the four datasets where we can control the intensities.
We select representative methods from standard PU learning (domain discriminator), propensity-weighting (SAR-EM), and graph PU learning (PU-GNN) to compare with RECO-SLIP.
We consider datasets with no shift (NS) where the source ratio of each category except the novel one is $0.5$.
We also consider datasets with minor shift (MS), which is a midpoint interpolation between shift (S) and no shift (NS).
The overall results are shown in Fig. \ref{fig:shift_intensity} and the performance rank of each method averaged over datasets under each shift intensity is presented in Table \ref{tab:intensity_rank}.
The advantage of RECO-SLIP diminishes as the shift intensity reduces.
However, it still ranks the highest under minor shift (MS) and the second highest under no shift (NS), surpassed by the state-of-the-art graph PU learning method, PU-GNN.
Overall, we can observe from Fig. \ref{fig:shift_intensity} and Table \ref{tab:intensity_rank} that RECO-SLIP is the most robust to subpopulation shifts and its overall performance rank is the highest.

\section{Conclusion}
In this study, we present RECO-SLIP, a new method for identifying nodes belonging to the novel categories in attributed graphs.
RECO-SLIP builds upon a recall-constrained learning framework to address subpopulation shifts and leverages a sample-efficient link prediction mechanism to preserve node subgroup structure.
Our experiment results demonstrate the superiority of RECO-SLIP over standard PU learning, propensity-weighting, and graph PU learning methods.
Furthermore, we conduct an ablation study and a shift intensity study, confirming the importance of selective link prediction and the robustness of RECO-SLIP across multiple shift intensities.
In terms of future work, novel node category detection under the shift in intra/inter-category edge connection probabilities is an exciting direction.
Through this extension, we will be able to capture a realistic scenario where the node interaction pattern changes from source to target, making novelty detection even more robust when deployed in the wild.

\appendix
%
%
%
%
%
%

%
%
%
\section{SCAR Does Not Hold Under Subpopulation Shift}\label{app:scar_break}
In the context of novel category detection, the selected completely at random (SCAR) assumption \cite{elkan2008learning} assumes each sample from a non-novel category has an equal probability of being in the source domain regardless of its data attributes. We show that SCAR does not hold under subpopulation shift.

As a reminder, $P_S$ and $P_{T,0}$ denote the source distribution and the non-novel distribution in the target.
They are mixtures of $K$ category distributions $G_1, G_2, \ldots, G_K$ with different mixture proportions defined by two probability vectors $\boldsymbol{\gamma}, \hat{\boldsymbol{\gamma}} \in \Delta^{K-1}$.
\begin{equation}
    P_S=\sum_{i=1}^{K} \gamma_{i} G_{i}, P_{T,0}=\sum_{i=1}^{K} \hat{\gamma}_{i} G_{i}, \boldsymbol{\gamma} \neq \hat{\boldsymbol{\gamma}}
\end{equation}

Let $S$ be a binary random variable denoting whether a sample belongs to the source domain or not, i.e. $S=1$ if the sample is in the source domain and $S=0$ otherwise.
Let $N$ be another binary random variable indicating if a sample is novel or not, i.e. $N=1$ if the sample is from the novel category and $N=0$ otherwise.
We use $\mathbf{X}$ and $\mathbf{x}$ to denote a random feature vector and its realization.
Let $p_S$ and $p_{T,0}$ be the probability density functions (PDF) associated with $P_S$ and $P_{T,0}$, respectively.
SCAR essentially assumes $P(S=1|N=0, \mathbf{X}=\mathbf{x}) = P(S=1|N=0)$.
However, we can show:

\begin{align*}
    & P(S=1|N=0, \mathbf{X}=\mathbf{x}) \\
 =& \frac{P(\mathbf{X}=\mathbf{x}|S=1, N=0)P(S=1, N=0)}{P(\mathbf{X}=\mathbf{x}|N=0)P(N=0)} \\
 =& \frac{P(\mathbf{X}=\mathbf{x}|S=1)}{P(\mathbf{X}=\mathbf{x}|N=0)} \frac{P(S=1, N=0)}{P(N=0)} \\
 =& \frac{P(\mathbf{X}=\mathbf{x}|S=1)}{P(S=0|N=0)P(\mathbf{X}=\mathbf{x}|N=0,S=0) + P(S=1|N=0)P(\mathbf{X}=\mathbf{x}|S=1)} \\
 & \hspace{55ex} \times \frac{P(S=1, N=0)}{P(N=0)} \\
 =& \frac{p_S(\mathbf{x})}{P(S=0|N=0)p_{T,0}(\mathbf{x}) + P(S=1|N=0)p_S(\mathbf{x})} \frac{P(S=1, N=0)}{P(N=0)} \\
 \neq& \frac{P(S=1, N=0)}{P(N=0)} \\
 =& P(S=1 | N=0)
\end{align*}
We get the first and final equalities through the Bayes rule.
The second equality holds because $N=0$ does not provide additional information when we already condition on $S=1$.
The third equality holds due to the law of total probability and the same reason that leads to the second equality.
We get the fourth equality by the definition of the PDFs.
Since $p_S(\mathbf{x}) \neq p_{T,0}(\mathbf{x})$, the inequality above holds unless $P(S=0|N=0)=0$, which is an extreme case that all non-novel samples are in the source domain and we do not consider such a case in our problem setting.
Therefore, we can conclude that SCAR does not hold under subpopulation shift.

\begin{table}[t]
\caption{Source ratio per category for three versions of Cora, CiteSeer, Computers, and Photo. Each entry represents the ratio of nodes belonging to a category that is assigned to the source domain. The source ratio of the last available category in each dataset is $0$ because the category is novel.}
\label{tab:src_ratio_all}
\begin{center}
\addtolength{\tabcolsep}{2pt}  
\scalebox{0.86}{
\begin{tabular}{lcccccccccc}
\toprule
& \multicolumn{10}{c}{\textbf{Category label}} \\
\midrule
                                   & 1          & 2          & 3          & 4          & 5          & 6          & 7    & 8    & 9    & 10 \\
\midrule
Cora-S                             & $0.1$      & $0.9$      & $0.1$      & $0.9$      & $0.1$      & $0.9$      & $0$   & -     & -    & -  \\
CiteSeer-S                         & $0.9$      & $0.1$      & $0.9$      & $0.1$      & $0.5$      & $0$        & -     & -     & -    & -  \\
Computers-S                        & $0.1$      & $0.9$      & $0.1$      & $0.9$      & $0.1$      & $0.9$      & $0.1$ & $0.9$ & $0.5$& $0$ \\
Photo-S                            & $0.9$      & $0.1$      & $0.9$      & $0.1$     & $0.9$      & $0.1$      & $0.5$ & $0$ & -    & - \\
\midrule
Cora-MS                             & $0.3$      & $0.7$      & $0.3$      & $0.7$      & $0.3$      & $0.7$      & $0$   & -     & -    & -  \\
CiteSeer-MS                         & $0.7$      & $0.3$      & $0.7$      & $0.3$      & $0.5$      & $0$        & -     & -     & -    & -  \\
Computers-MS                        & $0.3$      & $0.7$      & $0.3$      & $0.7$      & $0.3$      & $0.7$      & $0.3$ & $0.7$ & $0.5$& $0$ \\
Photo-MS                            & $0.7$      & $0.3$      & $0.7$      & $0.3$     & $0.7$      & $0.3$      & $0.5$ & $0$ & -    & - \\
\midrule
Cora-NS                             & $0.5$      & $0.5$      & $0.5$      & $0.5$      & $0.5$      & $0.5$      & $0$   & -     & -    & -  \\
CiteSeer-NS                         & $0.5$      & $0.5$      & $0.5$      & $0.5$      & $0.5$      & $0$        & -     & -     & -    & -  \\
Computers-NS                        & $0.5$      & $0.5$      & $0.5$      & $0.5$      & $0.5$      & $0.5$      & $0.5$ & $0.5$ & $0.5$& $0$ \\
Photo-NS                            & $0.5$      & $0.5$      & $0.5$      & $0.5$      & $0.5$      & $0.5$      & $0.5$ & $0$ & -    & - \\
\bottomrule
\end{tabular}}
\addtolength{\tabcolsep}{-2pt}
\end{center}
\end{table}

\section{Additional Dataset Details} \label{app:dataset}

\begin{figure}[t]
    \centering
    \includegraphics[width=0.9\textwidth]{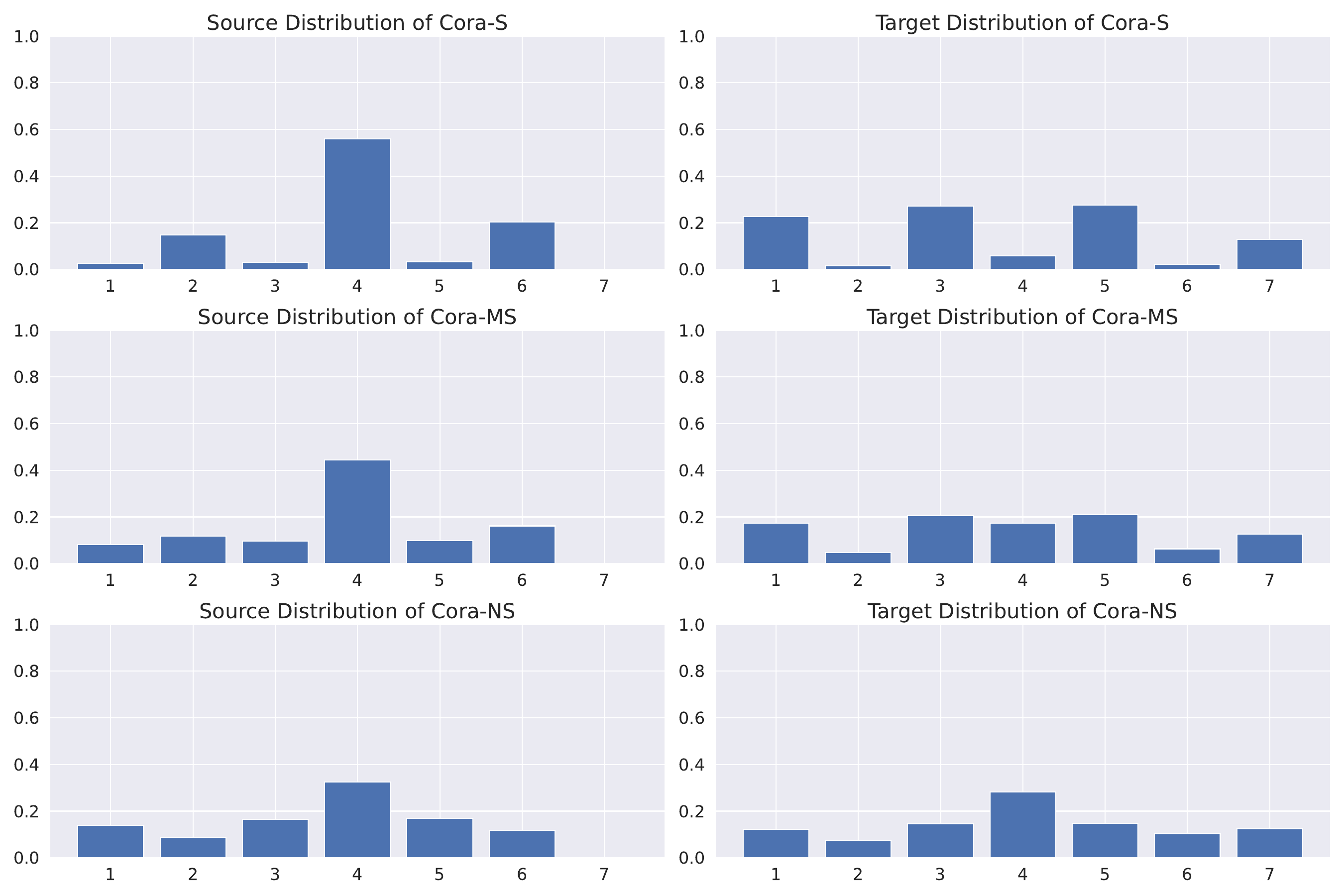}
    \caption{The source and target distributions of Cora-S, Cora-MS, and Cora-NS. The last category (Category 7) is the novel category so it does not show up in the source domain. The subpopulation shift exhibited in Cora-S is significant as more than half of the source nodes belong to Category 4 while Category 4 only takes up less than 10\% of the target nodes. In contrast, the relative proportions of the non-novel categories in the source and target are the same in Cora-NS.}
    \label{fig:cora_dist}
\end{figure}

\begin{figure}[h]
    \centering
    \includegraphics[width=0.9\textwidth]{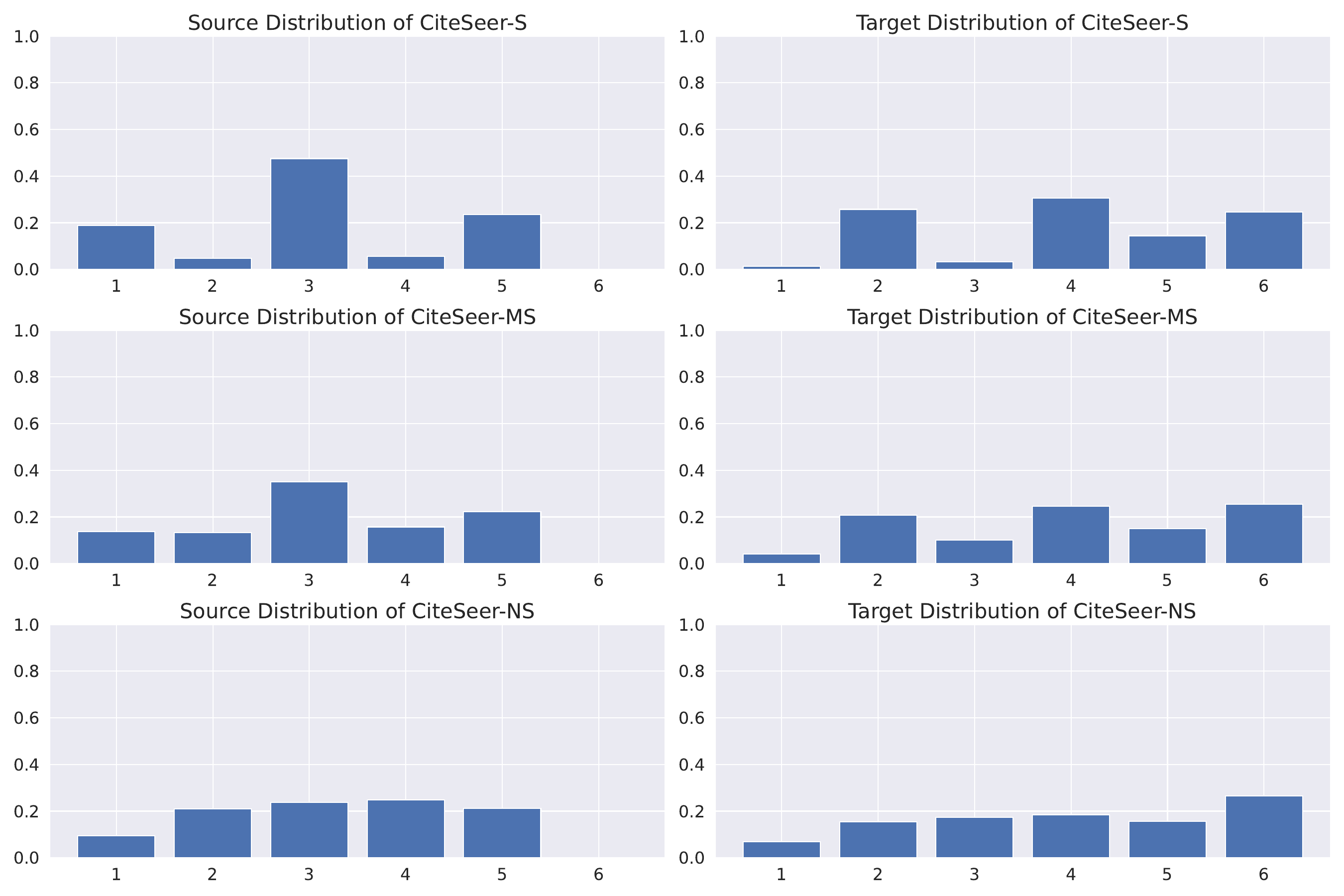}
    \caption{The source and target distributions of CiteSeer-S, CiteSeer-MS, and CiteSeer-NS.}
    \label{fig:citeseer_dist}
\end{figure}

\begin{figure}[h]
    \centering
    \includegraphics[width=\textwidth]{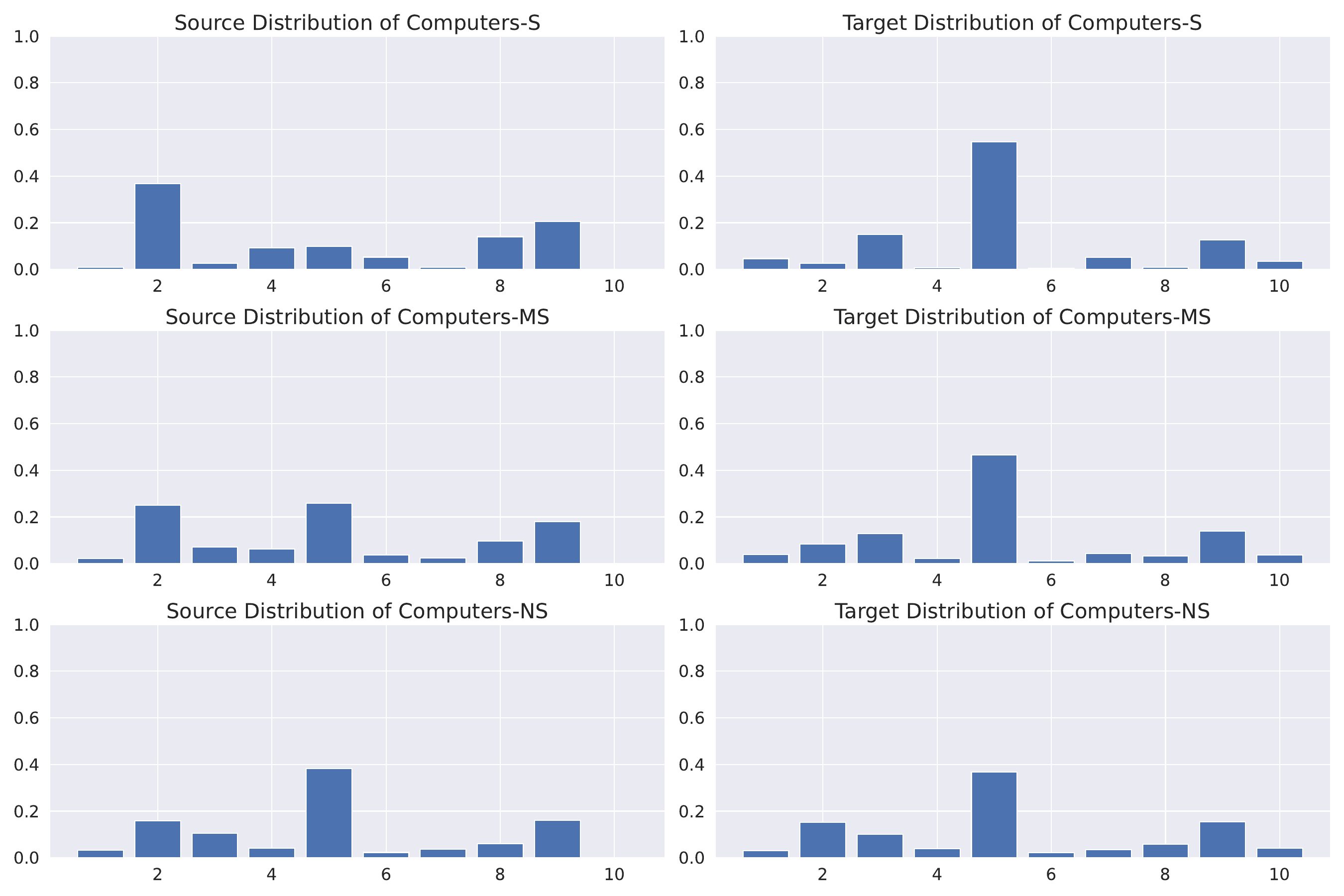}
    \caption{The source and target distributions of Computers-S, Computers-MS, and Computers-NS.}
    \label{fig:computers_dist}
\end{figure}

\begin{figure}[h]
    \centering
    \includegraphics[width=\textwidth]{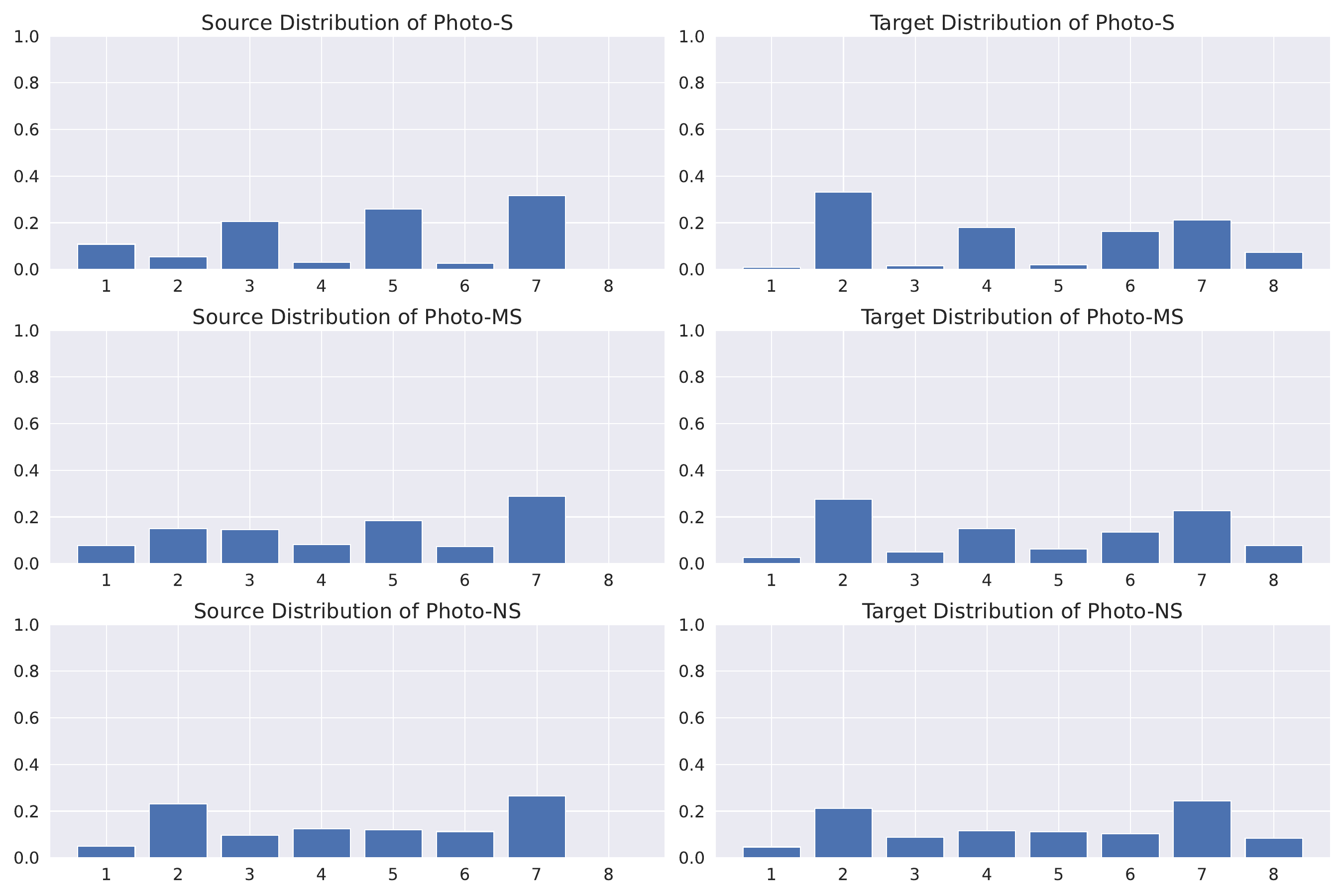}
    \caption{The source and target distributions of Photo-S, Photo-MS, and Photo-NS.}
    \label{fig:photo_dist}
\end{figure}

\begin{figure*}[h]
        \centering
        \begin{subfigure}[b]{0.46\textwidth}
            \centering
            \includegraphics[width=\textwidth]{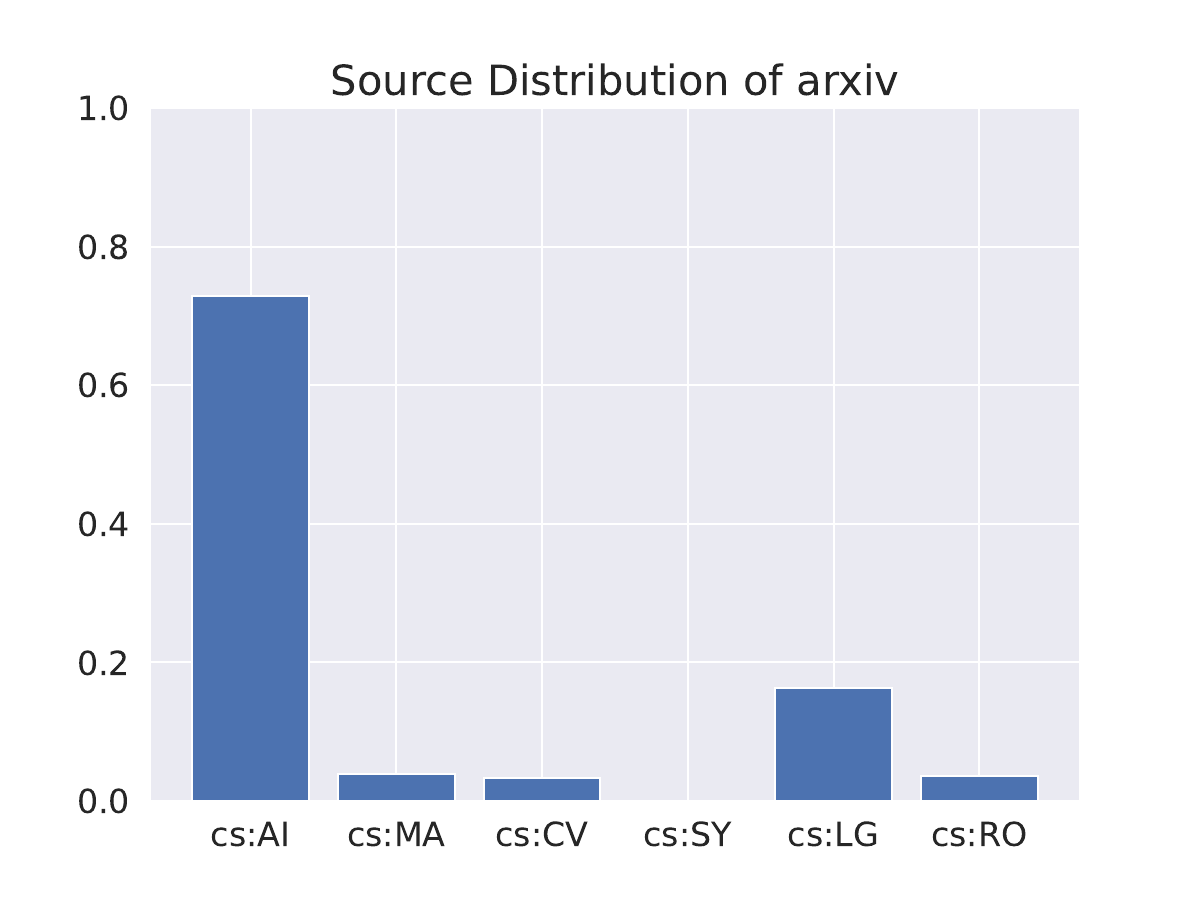}
            {{\small}}    
            \label{fig:arxiv_src_dist}
        \end{subfigure}
        \begin{subfigure}[b]{0.46\textwidth}  
            \centering 
            \includegraphics[width=\textwidth]{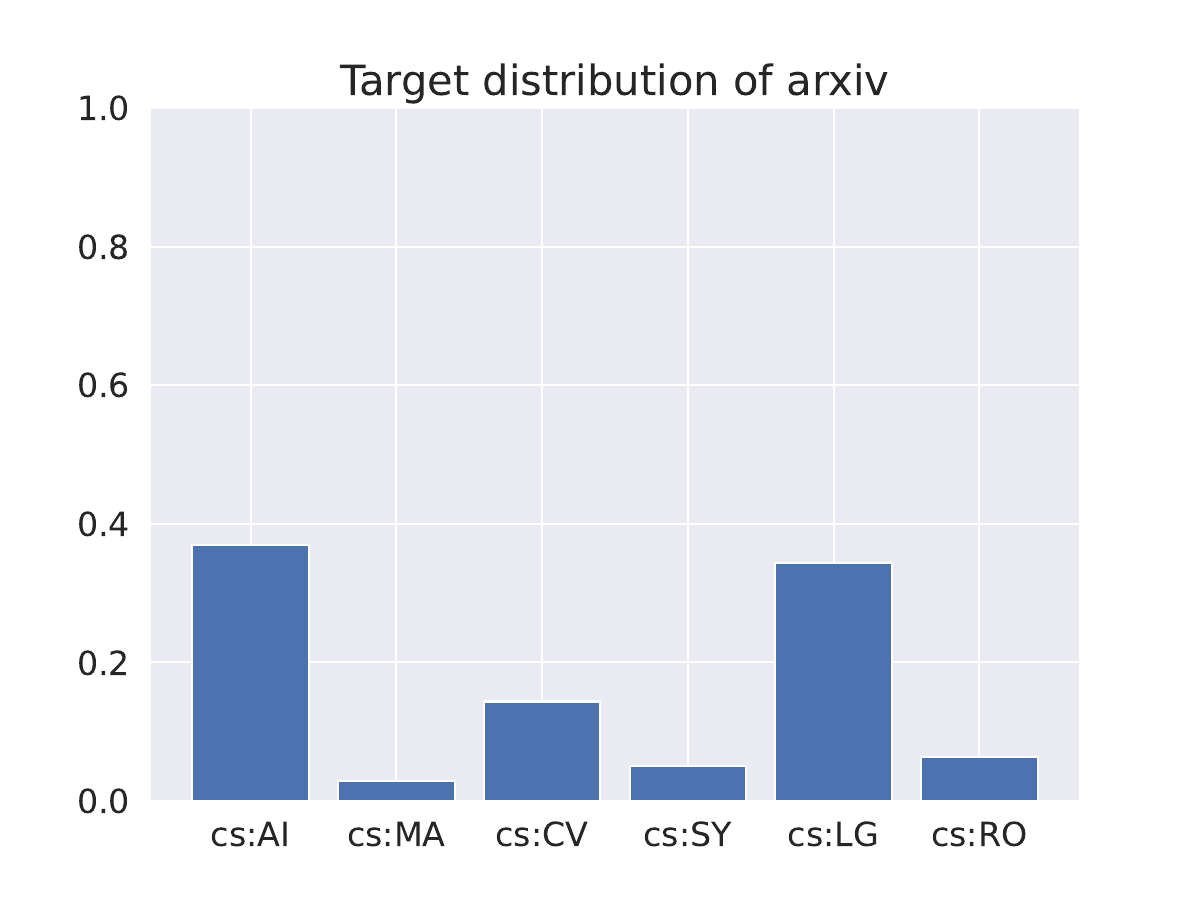}
            {{\small}}    
            \label{fig:arxiv_tgt_dist}
        \end{subfigure}
        \caption[]
        {\small A comparison between the source and target distributions of arxiv. The novel category cs:SY does not exist in the source but shows up in the target domain. The proportion of nodes falling in category cs:AI has a significant decrease from source to target while the ones of cs:CV and cs:LG increase notably.}
        \label{fig:arxiv_shift}
\end{figure*}

Cora \cite{sen2008collective} and CiteSeer \cite{sen2008collective} are academic citation graphs where nodes represent papers and edges represent citations.
Computers \cite{shchur2018pitfalls} and Photo \cite{shchur2018pitfalls} are product co-purchasing networks where nodes denote products and an edge between two products indicates that the two products were frequently bought together.
For our main experiments and study on shift intensity, we simulate 3 versions of these 4 datasets, denoted by the suffixes ``-S'', ``-MS'', and ``-NS'', to represent subpopulation shift intensities ranging from significant to nonexistent.
We show how we split nodes of each category to source and target for different versions of datasets in Table \ref{tab:src_ratio_all}.
We also plot their source and target distributions to visualize the subpopulation shifts in Fig. \ref{fig:cora_dist}, \ref{fig:citeseer_dist}, \ref{fig:computers_dist}, and \ref{fig:photo_dist}.
As for the train/validation/test split, we randomly select 80\% as train and 20\% as validation among the source and 60\% as train, 20\% as validation, and 20\% as test among the target.

arxiv is a subgraph of the ogbn-arxiv\footnote{\href{https://ogb.stanford.edu/docs/nodeprop/\#ogbn-arxiv}{https://ogb.stanford.edu/docs/nodeprop/\#ogbn-arxiv}} \cite{hu2020open} academic citation graph from $1990$ to $2012$ containing $6$ robotics-related categories: cs:AI (Artificial Intelligence), cs:MA (Multiagent Systems), cs:CV (Computer Vision and Pattern Recognition), cs:SY (Systems and Control), cs:LG (Machine Learning), and cs:RO (Robotics).
Nodes belonging to cs:SY did not exist until $2007$.
Therefore, we set the nodes with timestamps before $2007$ as the source and the rest as the target for our experiments.
The source and target distributions are visualized in Fig. \ref{fig:arxiv_shift}.
We select the nodes having timestamps in 2012 as the test set.
For both the source nodes and the target nodes excluding the test set, we randomly select 80\% as the training set and 20\% as the validation set.

\clearpage

\section{Model and Hyper-Parameter Configurations} \label{app:hyperparam}
\subsubsection{Model Structure}~
Models on all datasets have the following structure while \emph{gcn\_input\_dim}, \emph{gcn\_hidden\_dim}, \emph{gcn\_output\_dim}, \emph{mlp\_input\_dim}, and\\ \emph{mlp\_hidden\_dim} vary.
\begin{itemize}
    \item GCN encoder
    \begin{itemize}
        \item GCNConv(\emph{gcn\_input\_dim}, \emph{gcn\_hidden\_dim})
        \item BatchNorm1d
        \item ReLU
        \item Dropout(0.5)
        \item GCNConv(\emph{gcn\_hidden\_dim}, \emph{gcn\_output\_dim})
        \item BatchNorm1d
        \item ReLU
    \end{itemize}
    \item MLP head
    \begin{itemize}
        \item Linear(\emph{mlp\_input\_dim}, \emph{mlp\_hidden\_dim})
        \item BatchNorm1d
        \item ReLU
        \item Dropout(0.5)
        \item Linear(\emph{mlp\_hidden\_dim}, 2)
    \end{itemize}
\end{itemize}

\subsubsection{Random Seeds}~
We use the following 10 random seeds for all experiments: 10, 20, 30, 40, 50, 60, 70, 80, 90, 100.

\subsubsection{Optimizer}~
We use the Adam optimizer \cite{kingma2014adam} with a learning rate of $0.001$ for all methods, including both the primal and dual optimizers of RECO-SLIP.

\subsubsection{Mixture Proportion Estimation (MPE)}~
For PU learning methods that require MPE, we use the best bin estimator (BBE)\footnote{\href{https://github.com/acmi-lab/PU_learning/blob/main/estimator.py}{https://github.com/acmi-lab/PU\_learning/blob/main/estimator.py}} \cite{garg2021mixture} to estimate the prior.

\subsubsection{Domain Discriminator}
\begin{itemize}
    \item Maximum epochs: 2000
\end{itemize}
\subsubsection{uPU}
\begin{itemize}
    \item MPE epochs: 150
    \item Maximum total epochs: 1000
    \item Patience: 50
\end{itemize}
\subsubsection{nnPU}
\begin{itemize}
    \item MPE epochs: 150
    \item Maximum total epochs: 1000
    \item Patience: 50
\end{itemize}
\subsubsection{SAR-EM}
\begin{itemize}
    \item Maximum EM steps: 500
    \item Inner epochs for maximization (M): 200
    \item Patience for EM steps: 50
\end{itemize}
\subsubsection{LP-PUL}
\begin{itemize}
    \item Initial novel ratio: 0.5
    \item Label propagation layers: 3
    \item Label propagation $\alpha$: 0.9
\end{itemize}
\subsubsection{PU-GNN}
\begin{itemize}
    \item MPE epochs: 150
    \item Adapted Dist-PU $\delta$: 3
    \item Structural regularization weight: 0.1
    \item Maximum total epochs: 1000
    \item Patience: 50
\end{itemize}
\subsubsection{RECO-SLIP}
\begin{itemize}
    \item Link prediction loss weight $\xi$: 0.001
    \item $\Tilde{\boldsymbol{\alpha}}=[0.05, 0.1, 0.15, 0.2, 0.25]$
    \item Epochs: 1000
\end{itemize}

We list dataset-dependent hyper-parameters in Table \ref{tab:datasetdep_hyperparam}.
\begin{table}[t]
\caption{Dataset-dependent hyper-parameters}
\label{tab:datasetdep_hyperparam}
\begin{center}
\addtolength{\tabcolsep}{1pt}
\scalebox{0.9}{
\begin{tabular}{clccccc}
\toprule
&\multicolumn{1}{l}{\bf Hyper-parameters}  &\multicolumn{1}{c}{\bf Cora}  &\multicolumn{1}{c}{\bf CiteSeer}  &\multicolumn{1}{c}{\bf Computers}  &\multicolumn{1}{c}{\bf Photo} &\multicolumn{1}{c}{\bf arxiv}
\\ \midrule 
Model structure
& \emph{gcn\_input\_dim}                  & 1433                          & 3703                             & 767                               & 745                
            & 128 \\
& \emph{gcn\_hidden\_dim}                 & 16                            & 64                               & 16                                & 64                 
            & 64  \\
& \emph{gcn\_output\_dim}                 & 16                            & 32                               & 16                                & 32                 
            & 64 \\
& \emph{mlp\_input\_dim}                  & 16                            & 32                               & 16                                & 32                 
            & 64 \\
& \emph{mlp\_hidden\_dim}           & 8                             & 4                                & 8                                 & 32                 
            & 32 \\
\midrule
PU-GNN
& Structural regularization K             & 50                            & 50                               & 30                                & 30                 
            & 50 \\
\midrule
RECO-SLIP
& Initial dual variable $\lambda$         & 0.1                           & 0.1                              & 0.4                               & 0.1                 
            & 0.2 \\
& $\Tilde{\beta}$                         & 0.01                          & 0.05                             & 0.01                              & 0.05                
            & 0.01 \\
\bottomrule
\end{tabular}}
\addtolength{\tabcolsep}{-1pt}
\end{center}
\end{table}
\end{document}